\documentclass{article}
\usepackage[preprint,nonatbib]{nips_2018_mod}
\usepackage{graphicx}
\usepackage{amsmath}
\usepackage{amssymb}

\usepackage{hyperref}
\hypersetup{pagebackref=false,breaklinks=true,colorlinks,bookmarks=false,urlcolor=blue,linkcolor=blue,citecolor=blue}
\hypersetup{pdftitle={Surrogate Assisted Optimisation for Travelling Thief Problems}}
\hypersetup{pdfauthor={Majid Namazi, Conrad Sanderson, M.A. Hakim Newton, Abdul Sattar}}

\usepackage[british]{babel}
\usepackage[none]{hyphenat}
\makeatletter
\def\blfootnote{\xdef\@thefnmark{}\@footnotetext}
\makeatother

\begin{document}

\pagestyle{empty}

\title{Surrogate Assisted Optimisation\\for Travelling Thief Problems}
 
\author
  {
  Majid Namazi~\textsuperscript{$\dagger\ddagger$},
  Conrad Sanderson~\textsuperscript{$\dagger\ddagger$},
  M.A. Hakim Newton~\textsuperscript{$\ddagger$},
  Abdul Sattar~\textsuperscript{$\ddagger$}\\~\\
  {\small\textsuperscript{$\dagger$}~{Data61~/~CSIRO, Australia}}\\
  {\small\textsuperscript{$\ddagger$}~{Griffith University, Australia}}
  }

\maketitle

\begin{abstract}
\vspace{-1ex}
\small

The travelling thief problem (TTP) is a multi-component
optimisation problem involving two interdependent NP-hard components:
the travelling salesman problem (TSP)
and the knapsack problem (KP).
Recent state-of-the-art TTP solvers modify the underlying TSP and KP solutions in an iterative and interleaved fashion.
The TSP solution (cyclic tour) is typically changed in a deterministic way, 
while changes to the KP solution typically involve a random search,
effectively resulting in a quasi-meandering exploration of the TTP solution space.
Once a plateau is reached, the iterative search of the TTP solution space is restarted by using a new initial TSP tour.
We propose to make the search more efficient through an adaptive surrogate model 
(based on a customised form of Support \mbox{Vector} \mbox{Regression})
that learns the characteristics of initial TSP tours 
that lead to good TTP solutions.
The model is used to filter out non-promising initial TSP tours,
in effect reducing the amount of time spent to find a good TTP solution.
Experiments on a broad range of benchmark TTP instances
indicate that the proposed approach filters out a considerable number of non-promising initial tours,
at the cost of omitting only a small number of the best TTP solutions.

\end{abstract}

\section{Introduction}

Real-world optimisation problems composed of multiple
interdependent components are very challenging:
solving each component in isolation does not guarantee finding an 
optimal solution to the whole problem \cite{bonyadi2019evolutionary,michalewicz2012quo}. 
The travelling thief problem (TTP) combines
two interdependent components: the travelling salesman problem (TSP)
and the knapsack problem (KP), both NP-hard problems \cite{bonyadi2013travelling,polyakovskiy2014comprehensive}. 
In TTP, items are scattered among a set of cities;
a~thief goes on a cyclic tour through the cities
and collects a subset of the items into a rented knapsack.
As more items are collected, the speed of the thief decreases.
This increases the travelling time and hence the renting cost of the knapsack.
The aim in solving TTP is to maximise total gain
by simultaneously maximising the total profit of the collected items
and minimising the travelling time.
TTP can be viewed as a proxy for the arc-routing logistic problems
such as mail delivery, garbage collection, and network maintenance problems
where the order of visiting places or nodes is as important
as the length of the taken path \cite{corberan2015arc,mei2014improving}.

Recent state-of-the-art solvers for TTP \cite{el2018efficiently,namazi2019cooperative} 
solve the TSP and KP components in an iterative and interleaved fashion 
using a dedicated solver for each component.
The TSP solution (cyclic tour) is typically changed in a deterministic way, 
while changes to the KP solution (item collection plan) typically involve a random search.
This effectively results in a quasi-meandering exploration of the TTP solution space.
Upon reaching a plateau, the iterative search of the TTP solution space is restarted by employing a new initial TSP tour.
We have empirically observed that the final objective value does not vary
appreciably for similar initial cyclic tours,
suggesting that the overall search for a TTP solution with such solvers
involves redundant exploration of the solution space.
Furthermore, a subset of initial TSP tours (determined during the search)
will often lead to poor TTP solutions.

We propose to increase the efficiency of TTP search via filtering out non-promising initial TSP tours
through the use of an adaptive surrogate model.
Various surrogate models have been previously used to speed up computationally expensive simulations in fields such as groundwater modelling \cite{asher_2015}.
The proposed surrogate model approximates the final TTP objective value for any given initial TSP tour.
The model is built and automatically updated during the iterative search for a TTP solution.
It is based on non-linear Support Vector Regression \cite{smola2004tutorial}
with a novel kernel function
for measuring the similarity between TSP tours.
To our knowledge, this is the first time surrogate assisted optimisation
is used within the context of TTP.
Experiments on a wide subset of benchmark TTP instances
show that our proposed approach filters out a considerable number of the non-promising initial cyclic tours 
while missing only a small number of the best TTP solutions.

\section{Background}
\label{sec:background}

Each TTP instance has
a set {\small $\{1,\ldots,m\}$} of {$m$} items 
and 
a set {\small $\{1,\ldots,n\}$} of {$n$} cities.
The {distance} between each pair of cities {\small $i\ne i'$} is {\small $d(i,i') = d(i',i)$}.
Each {item} {$j$} is located at {city} {\small $l_j > 1$}.
Each item has {weight} {\small $w_j> 0$} and {profit} {\small $\pi_j > 0$}.

The thief starts a {cyclic tour} at city {$1$},
travels between cities (visiting each city once),
collects a subset of the items available in each city,
and returns to city {$1$}.
The tour is represented by using a permutation of {$n$} cities.
A given tour is represented as {$c$},
with {\small $c_k${\tiny~}={\tiny~}$i$} indicating that the {$k$}-th city in the tour {$c$} is {$i$},
and {\small $c(i)${\tiny~}={\tiny~}$k$} indicating that the position of city {$i$} in the tour {$c$} is {$k$}.
Here {\small $c_1${\tiny~}={\tiny~}$1$} and {\small $c(1)${\tiny~}={\tiny~}$1$}. 
A knapsack with a rent rate {$R$} per unit time and a weight capacity {$W$}
is rented by the thief to hold the collected items.
The {item collection plan} is represented by {\small $p$},
with {\small $p_i \in \{0,1\}$} indicating the collection state of item $i$.
An overall solution that provides a tour {$c$} and a collection plan {$p$} is expressed as {\small $\langle c,p \rangle$}.

The total weight of the items collected from city {$i$} is denoted by {\small $W_p(i)${\tiny~}={\tiny~}$\sum_{\forall l_j= i}w_jp_j$}.
The total weight of the items collected from the initial {$k$} cities in the tour {$c$}
is denoted by {\small $W_{c,p}(k)${\tiny~}={\tiny~}$\sum_{k'=1}^{k}W_p(c_{k'})$}.
The thief traverses from city {\small $c_k$} to the next city with speed {\small $v_{c,p}(k)$}.
The speed decreases as {\small $W_{c,p}(k)$} increases.
The speed at the city {\small $c_k$} is given by {\small $v_{c,p}(k)${\tiny~}={\tiny~}$v_\textrm{max} - W_{c,p}(k) \cdot (v_\textrm{max} - v_\textrm{min})/W$},
where {\small $v_\textrm{min}$} and {\small $v_\textrm{max}$} are minimum and maximum speeds, respectively.

Given a TTP solution {\small $\langle c,p \rangle$},
the total profit is {\small$P(p)${\tiny~}={\tiny~}$\sum_{i=1}^{m}p_i\pi_i$},
the travelling time to city {\small $c_k$} is {\small $T_{c,p}(k)${\tiny~}={\tiny~}$\sum_{k'=1}^{k-1}d(c_{k'},c_{k'+1})/v_{c,p}(k')$},
and the total travelling time is {\small $T(c,p)${\tiny~}={\tiny~}$T_{c,p}(n+1)${\tiny~}={\tiny~}$T_{c,p}(n)+d(c_n,c_1)/v_{c,p}(n)$}.
The goal of a TTP solution is to maximise the following {objective function}
over any viable {$c$} and {$p$}:
\begin{equation}
  G(c,p) = P(p) - R \cdot T(c,p)
  \label{eqn:ttp_objective_function}
\end{equation}%

\begin{figure}[!t]
  \centering
  \includegraphics[width=0.7\textwidth]{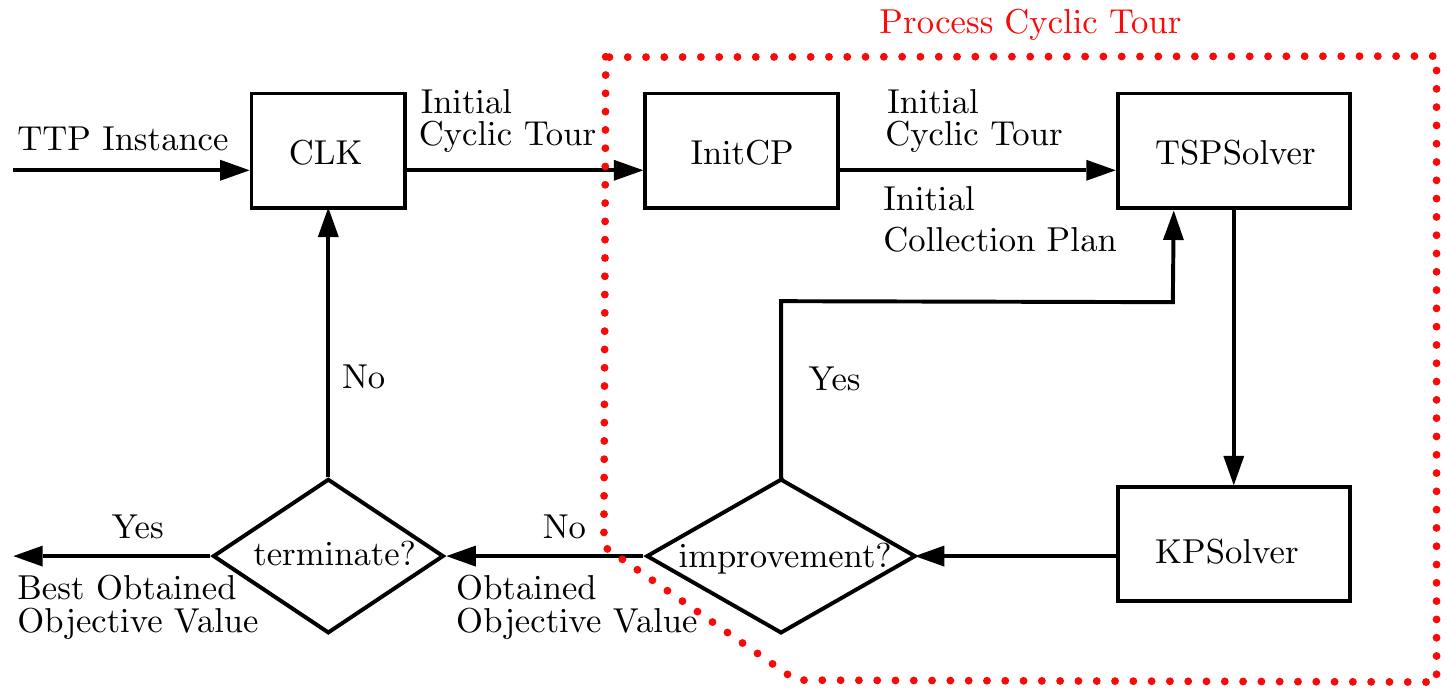}\\
  \caption{Structure of a typical restart-based cooperative solver for TTP.}
  \label{fig:cooperative}
\end{figure}

Recent solvers for TTP 
follow a cooperative strategy by solving the TSP and KP components in an interleaved fashion 
using a dedicated solver for each component
\cite{el2018efficiently,namazi2019socs,namazi2019cooperative}.
Fig.~\ref{fig:cooperative} shows 
how any given TTP instance is solved by these cooperative solvers.

For each given TTP instance, the Chained Lin-Kernighan (CLK) heuristic \cite{applegate2003chained} is used to generate an initial cyclic tour.
An initial collection plan is then obtained by a heuristic such as Insertion \cite{mei2014improving} or PackIterative \cite{faulkner2015approximate}. 
Next, the TSPSolver and KPSolver functions are invoked in an 
interleaved fashion to solve the TSP and KP components in successive rounds. 
In each iteration, in order to improve the objective value,
the TSPSolver deterministically chooses the best tour modifications,
while the KPSolver uses a stochastic local search
for improving the collection plan.

If the objective value is not improved in a round,
the solver restarts by asking the CLK routine to generate 
a new initial tour, provided that the termination condition is not met.
If the termination condition is met, 
the best obtained objective value and the corresponding solution are returned.

\begin{figure}[!t]
  \centering
  \includegraphics[width=0.7\textwidth]{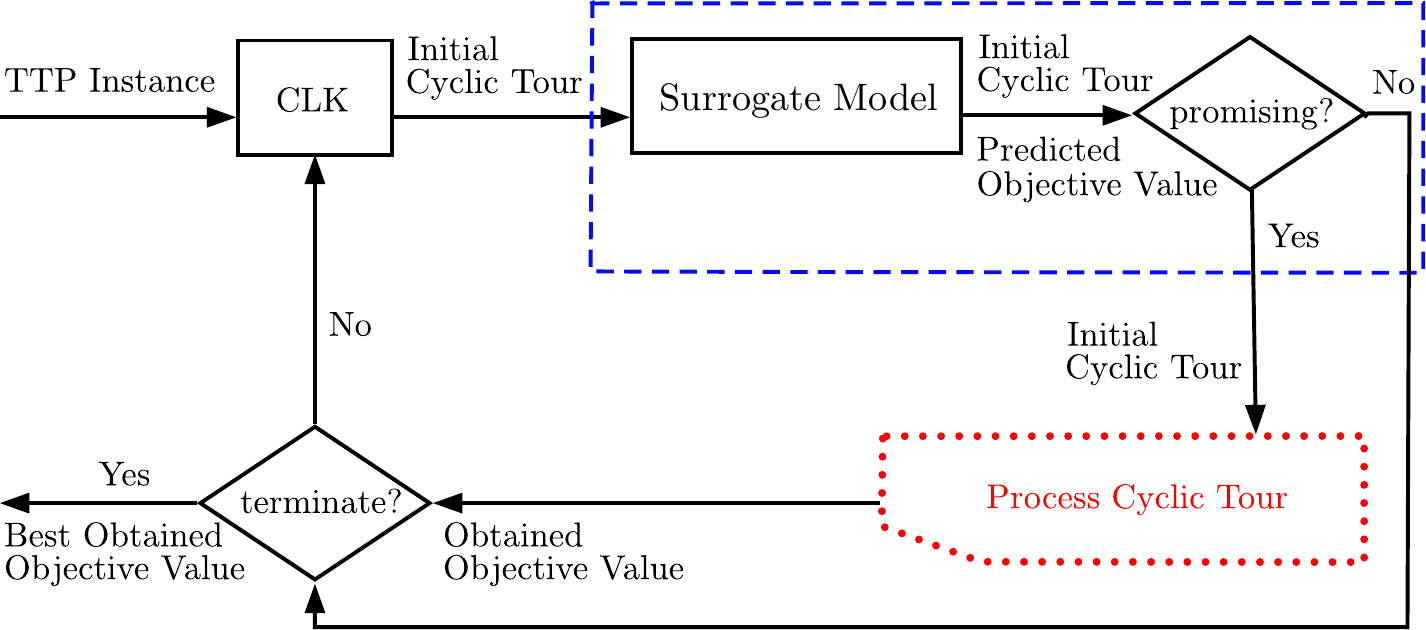}\\
  \caption{Addition of a surrogate model to the TTP solver in Fig.~\ref{fig:cooperative} in order to filter out non-promising initial tours.}
  \label{fig:coowithsur}
\end{figure}

\newpage
\section{Proposed Surrogate Model}
\label{sec:proposed}

Using the cooperative strategy shown in Fig.~\ref{fig:cooperative},
we have empirically observed that for similar initial cyclic tours,
the final objective value does not vary much.
This suggests that the overall search for a TTP solution with cooperative solvers
involves redundant exploration of the solution space.
Furthermore, a subset of initial TSP tours (determined during the search)
will often lead to poor TTP solutions.

Considering this semi-deterministic nature of the cooperative solvers,
we propose a surrogate model to emulate the set of functionality enclosed in the dotted red rectangle in Fig.~\ref{fig:cooperative}.
The surrogate model is used as shown in Fig.~\ref{fig:coowithsur} within the blue dashed rectangle.
For each generated initial tour, the surrogate model provides an approximation of the final TTP objective value.
If the generated initial tour appears non-promising,
it is disregarded from further optimisation (ie.,~filtered out).
Otherwise, the generated initial tour is allowed to proceed for further iterative optimisation.

\begin{figure}[b]
  \centering
  \includegraphics[width=0.4\columnwidth]{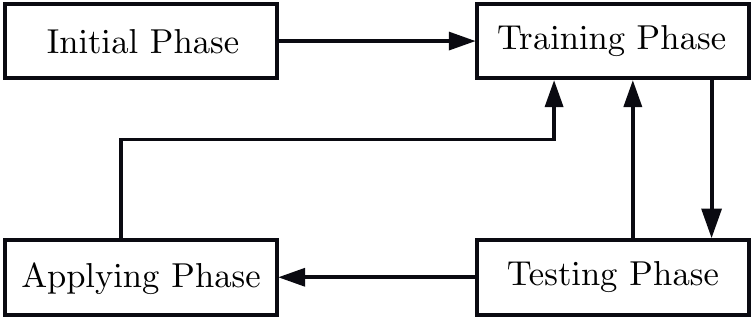}\\
  \caption{Transitions between phases of the surrogate model.}
  \label{fig:surgphases}
\end{figure}

For the surrogate model we propose an adaptive learning approach
employing non-linear kernel-based support vector regression (SVR) \cite{shawetaylor_2004,smola2004tutorial}.
While solving of a given TTP instance,
the surrogate model transitions between several phases as shown in Fig.~\ref{fig:surgphases}: 
initialisation, training, testing, and applying.
The phases and transitions between phases are elucidated below.

\subsection{\normalsize Initial Phase}

The given TTP instance is solved via restarting for a predefined number of times $t$,
where each run uses a new initial tour.
For any run $r$ in this phase,
the initial tour $c^r$ and the corresponding obtained final objective
value $g_r$ are kept as a pair $\langle c^r,g_r\rangle$ in a training set.

\subsection{\normalsize Training Phase}

To aid training the SVR, the training set is first normalised as follows.
Considering $g_\textrm{min}$ and $g_\textrm{max}$ as the minimum
and maximum objective values in the training set,
each $g_r$ is mapped to the [0,1] interval via:
\begin{equation}
n(g_r) =  \dfrac {g_{r} - g_\textrm{min}}  {g_\textrm{max} - g_\textrm{min}}
\end{equation}%

The resulting set 
{\small $X${\footnotesize =}$\{ \langle c^1,n(g_1)\rangle, \langle c^2,n(g_2)\rangle, ..., \langle c^t,n(g_t)\rangle \}$}
is used for training the kernel-based SVR.
Given a tour $c$,
SVR approximates the normalised final objective value via:
\begin{equation}
n(\widehat{g}) = \sum\nolimits_{r=1}^{t} (\alpha_r - \alpha_r^* ) \cdot k(c^r , c) + b
\end{equation}%

\noindent
where the SVR parameters $b$, $\alpha_r$ and $\alpha_r^*$ for $r \in [1,t]$
are computed as per \cite{shawetaylor_2004,smola2004tutorial}.
For the {kernel function} {\small $k(c^a,c^b)$}
we use a customised form of Gaussian radial basis function:
\begin{equation}
  k( c^a , c^b ) = \exp(-\gamma \cdot \Phi(c^a , c^b))
  \label{eqn:rbf_kernel}
\end{equation}%

\noindent
where $\gamma$ is a hyper-parameter,
while {\small $\Phi(c^a , c^b)$} is a measure of distance between tours $c^a$ and $c^b$
based on the positions of the cities in the tours:
\begin{equation}
\Phi(c^a , c^b) = \dfrac{1}{n}\cdot \sum\nolimits_{j=1}^{n} \dfrac{|c^a(j) - c^b(j)|}{n - 1}
\end{equation}%

\noindent
Here, {\small $c(j)$} indicates the position of city~$j$ in cyclic tour~$c$,
hence {\small $|c^a(j) - c^b(j)|$} is in the {\small $[0,n-1]$} range.
As such, {\small $\Phi(c^a , c^b)$} is in the {\small $[0,1]$} range.

The approximate final objective value {\small {\small $\widehat{g}$}} is obtained by 
reversing the normalisation:
\begin{equation}
\widehat{g} = n(\widehat{g}) \cdot (g_\textrm{max} - g_\textrm{min}) + g_\textrm{min}
\end{equation}%

\subsection{\normalsize Testing Phase}

Here the surrogate model is tested to ensure it has adequate accuracy and is retrained if required.
The given TTP instance is further solved using new initial tours for $\lambda \cdot t$ times,
where $t$ is the number of instances in the training set
and $\lambda$ is empirically selected as $0.20$.
In every run $r$,
for each generated initial cyclic tour $c$,
the actual final objective value $g$
as well as the approximate final objective value {\small $\widehat{g}$}
are obtained.

There are two conditions where retraining is triggered using an expanded training set.
Let us first define a Normalised Error (NE) measure as:
\begin{equation}
  \textrm{NE} = n(g) - n(\widehat{g})
  \label{eqn:ne}
\end{equation}%

\noindent
For any run which has {\small $\textrm{NE} > e$},
where $e$ is a predefined {error limit} empirically set to $0.02$,
the corresponding initial tour and actual final objective value are kept in a temporary buffer.
The temporary buffer is initialised to be empty at each start of the testing phase.

A form of moving cumulative average \cite{gama2010knowledge} of squares of all obtained \textrm{NE} values is kept,
referred to as {mean squared normalised error} ({MSNE}).
The MSNE is set to zero at each start of the testing phase.
For each run (with $r$ starting at~$1$),
{MSNE} is updated using:
\begin{equation}
  \textrm{MSNE}^\textrm{[new]} = \textrm{MSNE}^\textrm{[old]} + \dfrac{\textrm{NE}^2-\textrm{MSNE}^\textrm{[old]}}{r}
  \label{eqn:msne_update}
\end{equation}%

The first condition for retraining is as follows.
If a run is encountered that has {\small $g < g_\textrm{min}$} or {\small $g>g_\textrm{max}$},
the corresponding initial tour and final objective value are added to the temporary buffer,
followed by incorporating the buffer into the training set
and immediately restarting the training phase.

The second condition is as follows.
If {\small $\textrm{MSNE} > e$} after processing all $\lambda \cdot t$ initial tours,
the temporary buffer is incorporated into the training set
and the training phase is restarted.

\subsection{\normalsize Applying Phase}

Here the surrogate model is employed for filtering out (disregarding) non-promising initial tours.
Retraining may also be triggered in a similar manner to the testing phase.

\begin{figure}[!h]
  \centering
  \includegraphics[width=0.6\columnwidth]{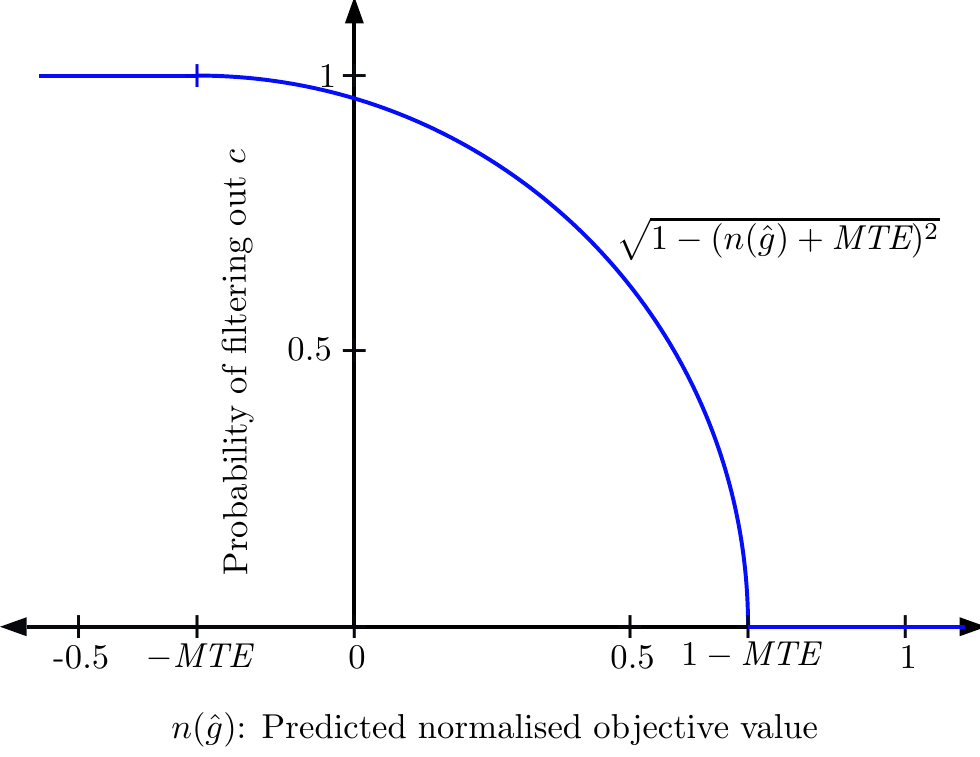}\\
  \caption{Probability of filtering out tour $c$ based on its approx.~normalised final objective value~{\small $n(\widehat{g})$}.}
  \label{fig:IgnGiagram}
\end{figure}

We define {maximum tolerable error} ({MTE}) as:
\begin{equation}
  \textrm{MTE} = \beta \cdot \sqrt{\textrm{MSNE}}
  \label{eqn:MTE}
\end{equation}%

\noindent
where $\beta$ is a hyper-parameter.
For a given initial tour $c$, the corresponding approximate normalised final objective value {\small $n(\widehat{g})$} is obtained.
If {\small $n(\widehat{g}) \geq 1 - \textrm{MTE}$},
the tour $c$  is allowed to proceed for further iterative optimisation.
Otherwise, the tour is filtered out
either when {\small $n(\widehat{g}) < -\textrm{MTE}$},
or with a probability of {\small $\sqrt{1 - (n(\widehat{g})+\textrm{MTE})^2}$}.
Fig.~\ref{fig:IgnGiagram} shows how the probability of filtering out $c$ is based on the value of {\small $n(\widehat{g})$}.

Whenever an initial tour $c$ is not filtered out,
the given TTP instance is solved using $c$ and the actual final objective value $g$ is obtained.
The corresponding {NE} is computed as per Eqn.~(\ref{eqn:ne}),
followed by updating {MSNE} as per Eqn.~(\ref{eqn:msne_update}).

Similar to the testing phase,
if {\small $\textrm{NE} > e$},
the corresponding initial tour and actual final objective value are stored in the temporary buffer initialised in the preceding testing phase.
If {\small $g < g_\textrm{min}$} or {\small $g > g_\textrm{max}$},
the corresponding initial tour and final objective value are added to the temporary buffer,
followed by incorporating the buffer into the training set
and immediately restarting the training phase.

Furthermore, retraining occurs whenever
{\small $\textrm{MSNE} > e$}
or
the number of runs with \mbox{\small $\textrm{NE} > e$} exceeds
{\small $\frac{1}{2} |X| $},
where {\small $|X|$} is the current cardinality of the training set.
This approach aims to increase the size of the training set during the early stages of optimisation,
while reducing the likelihood of retraining on large sets during later stages.

The rationale behind the probabilistic method to filter out non-promising initial tours is twofold.
(1) There is always a chance of under-prediction of the final objective value,
especially for (desirable) large final objective values.
(2) Not filtering out tours with small predicted final objective values
makes the updated {MSNE} value more accurate over the runs in this phase.

\newpage
\section{Experiments}
\label{sec:experiments}

As a baseline TTP solver we use the recently proposed cooperative coordination (CoCo) solver \cite{namazi2019cooperative}.
We extend the solver with the proposed surrogate model and refer to it as CoCo-SM.

We use a broad subset of medium and large-sized benchmark TTP instances
introduced by \cite{polyakovskiy2014comprehensive}. 
Considered instances are placed into 3 categories.
Each category has 32 instances with a range of 574 to 7397 cities.
In category~A, there is only one item in each city; the profits and weights of the items are strongly correlated; knapsack capacity is relatively small.
In category~B, there are 5 items in each city; the profits and weights of the items are uncorrelated; the weights of the items are similar to each other; knapsack capacity is moderate.
In category~C, there are 10 items in each city; the profits and weights of the items are uncorrelated; knapsack capacity is high.

Experiments were performed with {\small $\beta \in \{ 0, 1, 2, 3 \}$}
for computing MTE in Eqn.~(\ref{eqn:MTE}).
Both solvers were run on each TTP instance 10 times. In each run, CoCo solver was initially run for 1000 restarts on each instance.
CoCo-SM was then run on the same instance using the same set of 1000 initial tours generated and used by CoCo for that instance.
As such, we can see the effects if the CoCo-SM solver was used instead of the CoCo solver using the same set of the initial tours.

The initial tours and the corresponding actual final objective values in the first 10\% of the restarts in each run on each instance
were used to build the initial surrogate model in CoCo-SM.
For the custom RBF kernel in Eqn.~(\ref{eqn:rbf_kernel}), the hyper-parameter {\small $\gamma$} was set to $1$ based on preliminary experiments.

Table~\ref{tab:conf2r} shows the results
with the configuration of $\beta = 2$ in Eqn.~(\ref{eqn:MTE}).
The results are presented as the percentage of filtered out tours
and the corresponding number of missed best solutions (out of 10 runs).
In a ``missed best solution'', 
an initial tour that led to the best possible solution in a run
is incorrectly filtered out. 
The higher the percentage of filtered out tours, the better.
The lower number of missed best solutions, the better.
The results show that on average about 30\% of initial tours are filtered out
at the cost of missing about 1 best solution out of 10.

The overhead for training and using the surrogate model is overall negligible.
For example, for the hardest to solve instance (the last instance in category C),
around 10,000 seconds are required to process 1000 initial tours by the CoCo solver,
while about 15 seconds are required to train and use the surrogate model during processing of all the tours in CoCo-SM.
As such, if 30\% of the initial tours are filtered out,
the solver requires about 30\% less time to solve a given TTP instance.

\begin{figure*}[!b]
  \centering
  \hfill
  \begin{minipage}{0.325\textwidth}
    \centering
    \includegraphics[width=1.025\textwidth]{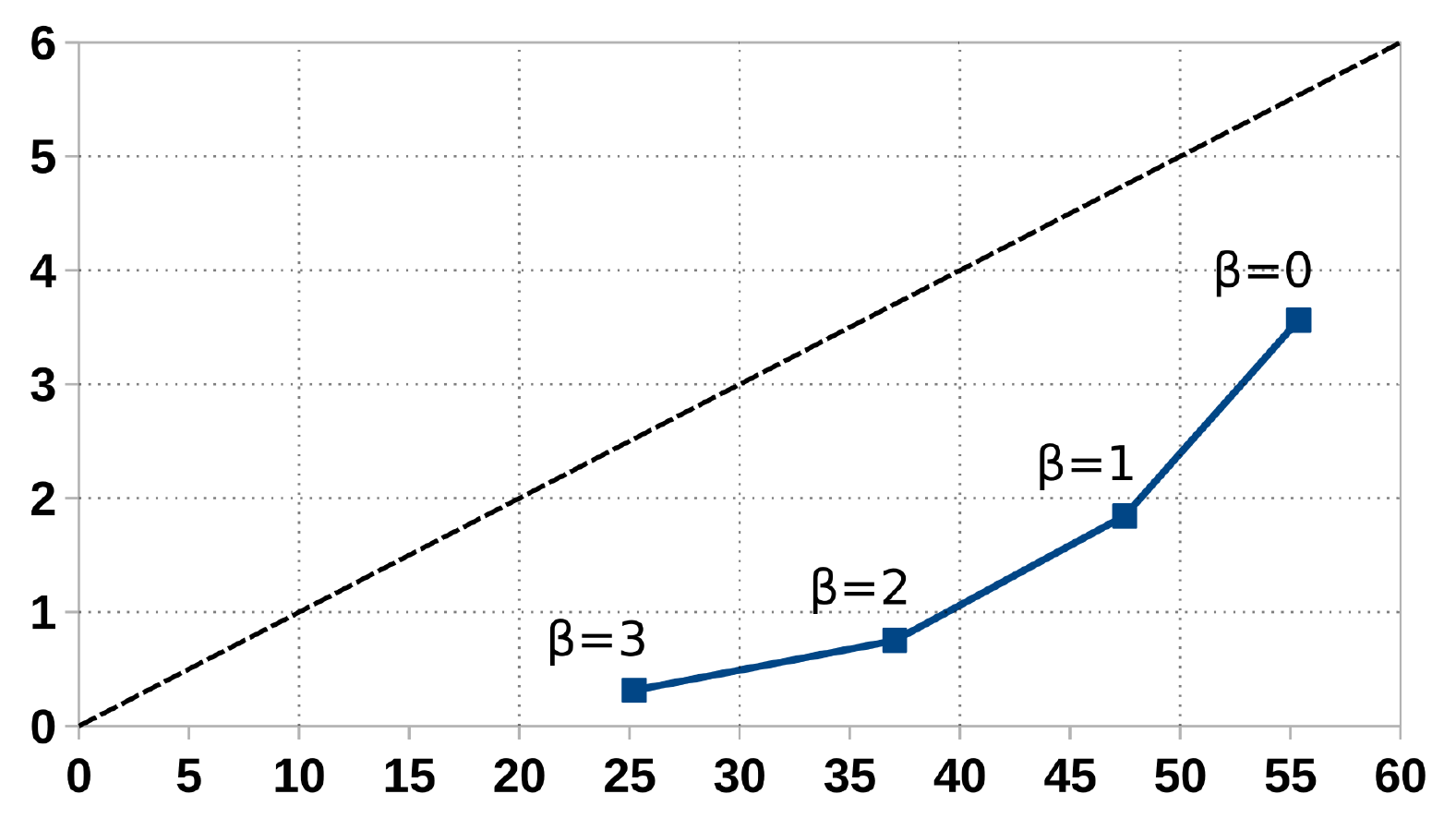}\\
    {\small TTP Category A}
  \end{minipage}
  \hfill
  \begin{minipage}{0.325\textwidth}
    \centering
    \includegraphics[width=1.025\textwidth]{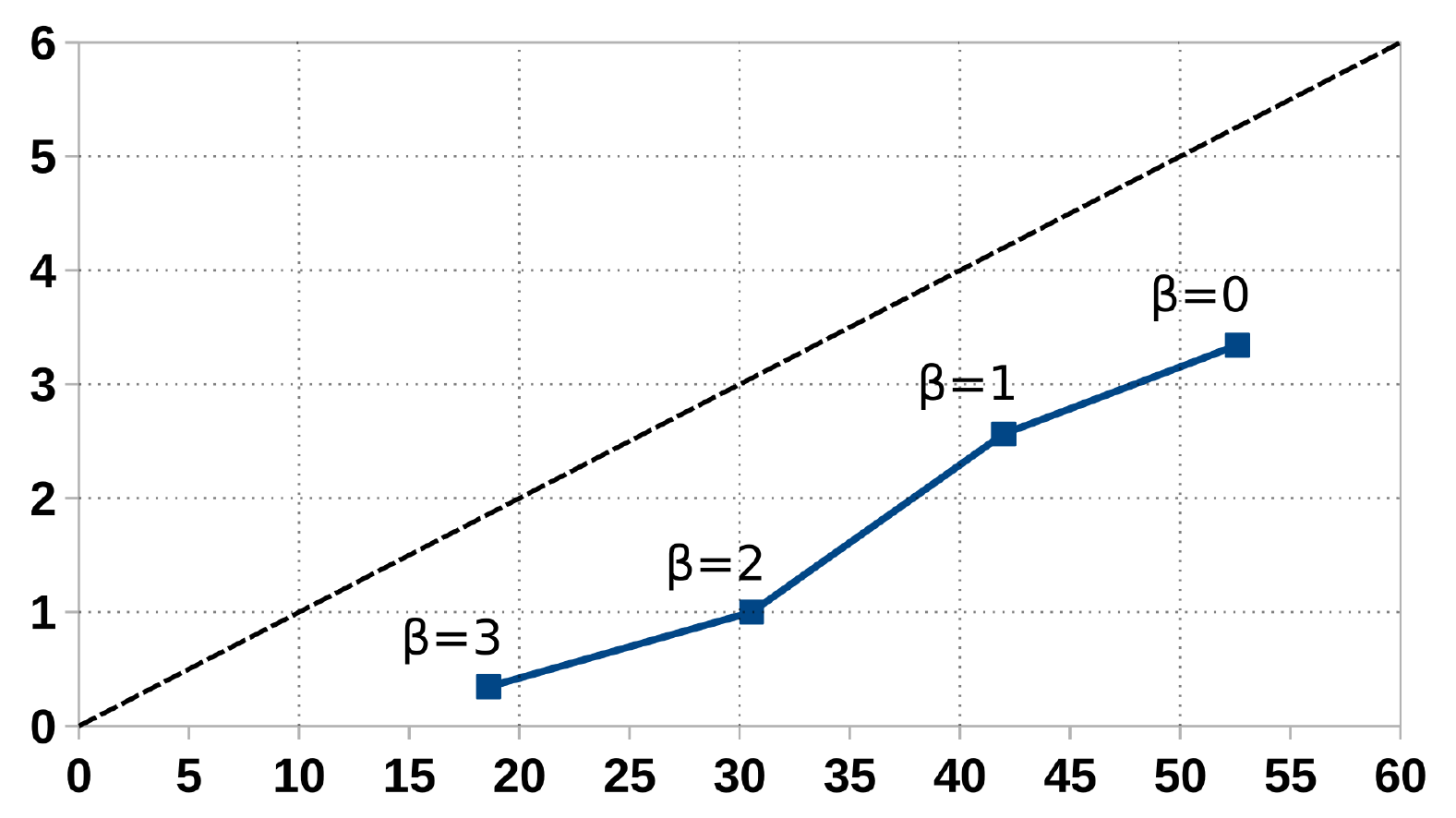}\\
    {\small TTP Category B}
  \end{minipage}
  \hfill
  \begin{minipage}{0.325\textwidth}
    \centering
    \includegraphics[width=1.025\textwidth]{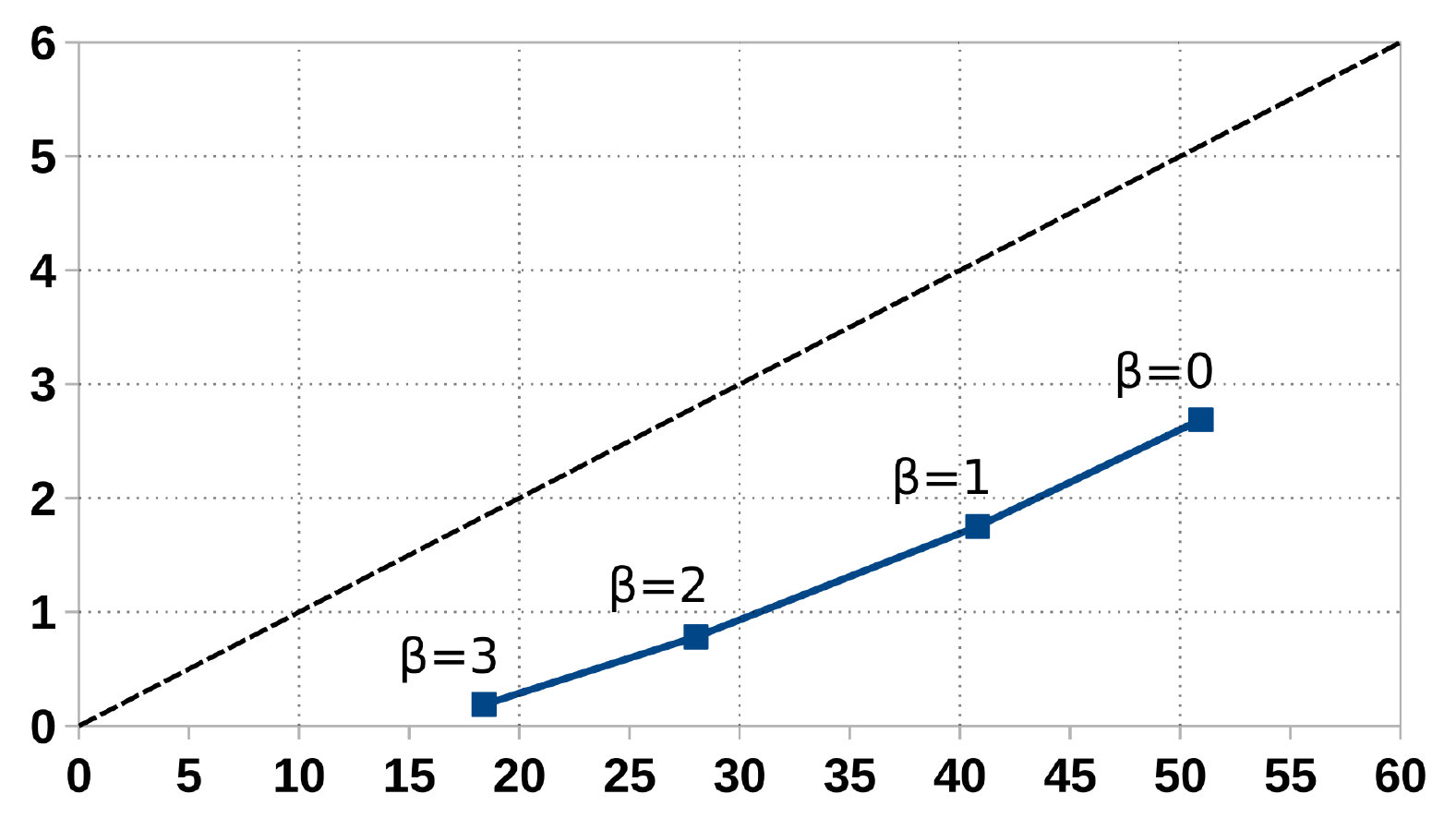}\\
    {\small TTP Category C}
  \end{minipage}
    \caption{%
      {\it x}-axis: average percentage of the filtered out initial cyclic tours;
      {\it y}-axis: average number of the missed best solutions (out of 10). 
      Each distinct point corresponds to a configuration of {\small $\beta \cdot \sqrt{\textrm{MSNE}}$}
      as defined in Eqn.~(\ref{eqn:MTE}).
      The dashed diagonal line represents the number of expected missed best solutions
      when random filtering is used instead of filtering based on the surrogate model.
      Better performance is indicated by an operating point that is further away
      from the diagonal line, moving towards the bottom right corner.
    }
  \label{ScPlots}
\end{figure*}

Fig.~\ref{ScPlots} shows the results for {\small $\beta \in \{ 0, 1, 2, 3 \}$} in Eqn.~(\ref{eqn:MTE}),
where the the average number of the missed best solutions 
is plotted against the average percentage of filtered out initial tours.
The dashed diagonal line represents the number of expected missed best solutions
when random filtering is used instead of filtering based on the surrogate model.
As such, better performance is indicated by an operating point that is further away
from the diagonal line, moving towards the bottom right corner.

The results indicate that the proposed surrogate model 
achieves considerably better filtering than simple random filtering.
The results also show that there is a trade-off:
the larger the percentage of filtered out tours,
the higher the chance of missing the best solution.

\begin{table}[!t]
\centering
\footnotesize
\begin{tabular}{|l|r|r|r||r|r|r|}
\hline
 & \multicolumn{3}{c||}{\textbf{\% of filtered out tours}} & \multicolumn{3}{l|}{\textbf{num. missed best sol.}} \\ \hline
\textbf{Instance} & \multicolumn{1}{c|}{\textbf{~~~A~~~}} & \multicolumn{1}{c|}{\textbf{~~~B~~~}} & \multicolumn{1}{c||}{\textbf{~~~C~~~}} & \multicolumn{1}{c|}{\textbf{~~~A~~~}} & \multicolumn{1}{l|}{\textbf{~~~B~~~}} & \multicolumn{1}{c|}{\textbf{~~~C~~~}} \\ \hline
u574 & 36.7 & 42.8 & 33.5 & 0 & 0 & 0 \\
rat575 & 33.4 & 32.4 & 37.7 & 1 & 3 & 1 \\
p654 & 41.5 & 25.7 & 41.0 & 1 & 2 & 3 \\ 
d657 & 25.8 & 31.0 & 20.9 & 2 & 1 & 0 \\ \hline
u724 & 33.0 & 31.4 & 33.7 & 2 & 1 & 1 \\
rat783 & 53.0 & 37.4 & 27.3 & 0 & 1 & 0 \\
dsj1000 & 88.0 & 51.4 & 30.1 & 0 & 2 & 1 \\
pr1002 & 20.9 & 55.6 & 41.1 & 0 & 3 & 4 \\ \hline
u1060 & 36.2 & 47.0 & 42.6 & 0 & 0 & 0 \\
vm1084 & 48.3 & 39.8 & 30.5 & 0 & 1 & 0 \\
pcb1173 & 42.0 & 30.5 & 32.7 & 0 & 0 & 0 \\
d1291 & 25.2 & 30.8 & 33.4 & 0 & 0 & 0 \\ \hline
rl1304 & 57.6 & 41.0 & 43.9 & 0 & 0 & 0 \\
rl1323 & 51.6 & 33.9 & 40.1 & 0 & 1 & 1 \\
nrw1379 & 30.8 & 14.8 & 13.6 & 1 & 0 & 0 \\
fl1400 & 41.9 & 56.6 & 58.3 & 0 & 2 & 0 \\ \hline
u1432 & 26.7 & 14.8 & 10.6 & 1 & 1 & 2 \\ 
fl1577 & 43.5 & 30.7 & 29.5 & 1 & 0 & 0 \\ 
d1655 & 41.4 & 22.5 & 24.2 & 2 & 0 & 0 \\
vm1748 & 27.9 & 41.3 & 33.9 & 0 & 0 & 2 \\ \hline
u1817 & 38.8 & 13.6 & 6.2 & 2 & 0 & 0 \\ 
rl1889 & 39.1 & 22.6 & 28.3 & 0 & 0 & 2 \\ 
d2103 & 44.5 & 57.4 & 52.0 & 4 & 4 & 1 \\
u2152 & 25.7 & 15.4 & 17.2 & 0 & 0 & 0 \\ \hline
u2319 & 30.4 & 28.4 & 23.3 & 1 & 4 & 0 \\ 
pr2392 & 40.6 & 18.1 & 17.3 & 0 & 1 & 1 \\ 
pcb3038 & 26.9 & 15.6 & 13.4 & 1 & 0 & 2 \\
fl3795 & 31.4 & 13.2 & 12.5 & 0 & 0 & 0 \\ \hline
fnl4461 & 20.9 & 11.6 & 6.1 & 3 & 1 & 0 \\ 
rl5915 & 25.8 & 24.9 & 18.7 & 1 & 2 & 2 \\ 
rl5934 & 29.7 & 35.2 & 30.8 & 1 & 1 & 2 \\
pla7397 & 26.3 & 9.9 & 12.3 & 0 & 1 & 0 \\ \hline\hline
{\bf Average} & 37.0 & 30.5 & 28.0 & 0.75 & 1 & 0.78 \\ \hline
\end{tabular}
\vspace{1ex}
\caption
  {
  Average percentage of the filtered out initial cyclic tours 
  with the corresponding number of missed best solutions (out of 10 runs)
  using the \mbox{CoCo-SM} solver
  with the configuration of $\beta = 2$ in Eqn.~(\ref{eqn:MTE}).
  Three~categories of TTP instances are used.
  Category~A:
    knapsack capacity is relatively small;
    1~item in each city;
    weights and profits of items are highly correlated.
  Category~B:
    knapsack capacity is moderate;
    5~items in each city;
    weights and profits of items are uncorrelated;
    weights of all items are similar.
  Category~C: 
    knapsack capacity is high;
    10~items in each city;
    weights and profits of items are uncorrelated.
  }
\label{tab:conf2r}
\end{table}

\newpage
\section{Conclusion}
\label{sec:conclusion}
\vspace{-1ex}

We have proposed to increase the efficiency of recent TTP solvers
by incorporating a surrogate model that assists in pruning the starting points
for restart-based optimisation.

In recent TTP solvers, the solutions to the underlying TSP and KP problems are changed in an iterative and interleaved fashion.
The TSP solution (cyclic tour) is typically changed in a deterministic way, 
while changes to the KP solution typically involve a random search,
resulting in a quasi-meandering exploration of the TTP solution space.
Upon reaching a plateau, the iterative search of the TTP solution space is restarted by employing a new initial TSP tour.

The proposed surrogate model,
based on Support Vector Regression with a novel kernel,
adaptively learns the characteristics of initial TSP tours that lead to good TTP solutions.
Non-promising initial TSP tours are detected and disregarded,
in effect reducing the amount of time spent to find a good TTP solution.

Experiments on benchmark TTP instances
show that the proposed approach removes a considerable number of non-promising initial tours,
at the cost of missing a small number of the best TTP solutions.


\bibliographystyle{abbrv}
\bibliography{references}

\end{document}